\documentclass{article} 
\usepackage{iclr2015,times}
\usepackage{hyperref}
\usepackage{mystuff-nlp}
\usepackage{tikz}
\usepackage{url}
\usepackage{latexsym}
\usepackage{color}
\usepackage{float}
\usepackage{amsmath,amsfonts,amssymb}
\usepackage{algorithm}
\usepackage{algpseudocode}
\usepackage{graphicx,subfigure}
\usepackage{booktabs}
\usepackage{multirow}

\newcommand{\vu}{\mathbf{u}}
\newcommand{\vv}{\mathbf{v}}

\newcommand{\sysname}{\textsc{Fema}}

\newcommand{\example}[1]{`#1'}

\title{Unsupervised Domain Adaptation with Feature Embeddings}

\author{
Yi Yang \& Jacob Eisenstein \\
School of Interactive Computing\\
Georgia Institute of Technology\\
\texttt{\{yiyang, jacobe\}@gatech.edu} \\
}

\begin{document}

\maketitle

\begin{abstract}
Representation learning is the dominant technique for unsupervised domain adaptation, but existing approaches often require the specification of ``pivot features'' that generalize across domains, which are selected by task-specific heuristics. We show that a novel but simple \emph{feature embedding} approach provides better performance, by exploiting the feature template structure common in NLP problems. 

\end{abstract}

\section{Introduction}
\label{sec:intro}
Domain adaptation is crucial if natural language processing is to be successfully employed in high-impact application areas such as social media, patient medical records, and historical texts. Unsupervised domain adaptation is particularly appealing, since it requires no labeled data in the target domain. Some of the most successful approaches to unsupervised domain adaptation are based on representation learning: transforming sparse high-dimensional surface features into dense vector representations, which are often more robust to domain shift~\citep{blitzer2006domain,glorot2011domain}. However, these methods are computationally expensive, and often require special task-specific heuristics to select good ``pivot features''.

%

We present \sysname~(\textbf{F}eature \textbf{EM}beddings for domain \textbf{A}daptation), a novel representation learning approach for domain adaptation in structured feature spaces. Like prior work in representation learning, \sysname~learns dense features that are more robust to domain shift. However, \sysname~diverges from previous approaches based on reconstructing pivot features; instead, it uses techniques from neural language models to directly obtain low-dimensional embeddings. \sysname~outperforms prior work on adapting POS tagging from the Penn Treebank to web text.

\section{Learning feature embeddings}
\label{sec:embed}

\begin{figure*}
\centering
\includegraphics[scale=.45]{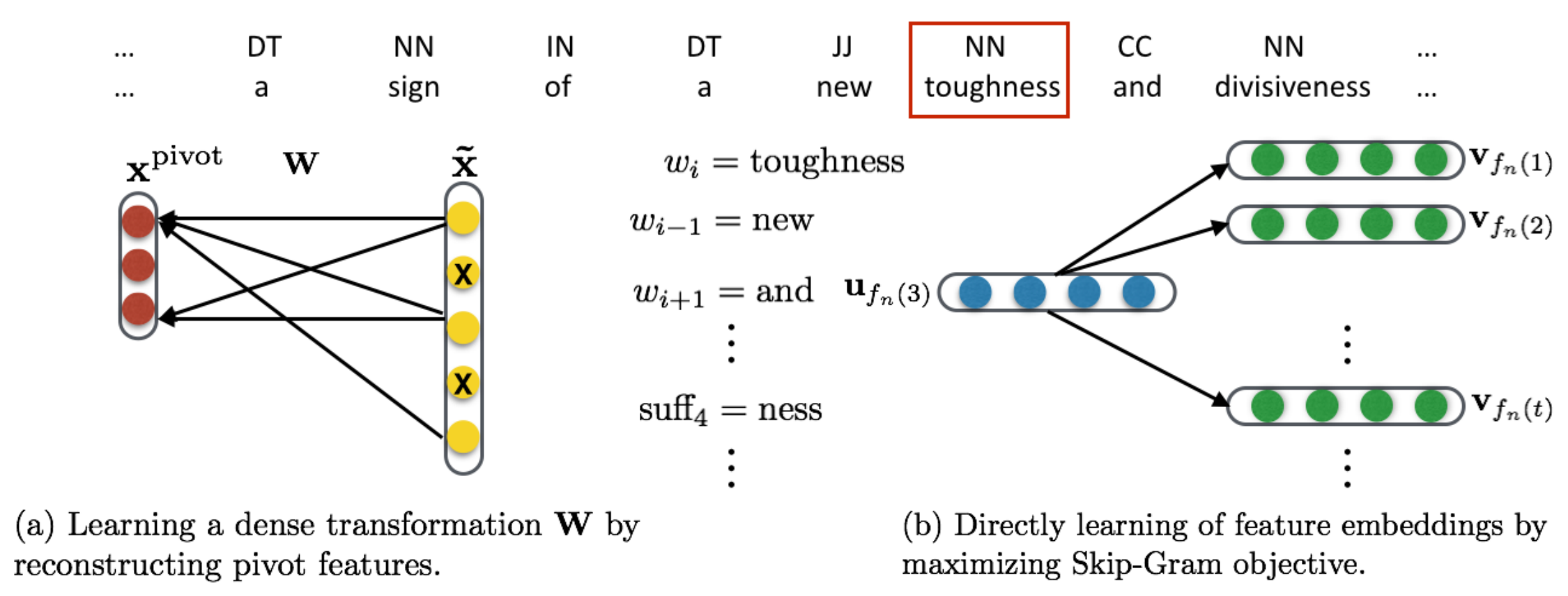}
\caption{\label{fig:dacomp} Representation learning techniques in structured feature spaces}
\end{figure*}

Feature co-occurrence statistics are the primary source of information driving many unsupervised methods for domain adaptation. For example, both Structural Correspondence Learning (SCL; Blitzer et al., 2006)\nocite{blitzer2006domain} and Denoising Autoencoders~\citep{chen2012marginalized} learn to reconstruct a subset of ``pivot features'', as shown in Figure~\ref{fig:dacomp}(a). The reconstruction function is then employed to project each instance into a dense representation, which will hopefully be better suited to cross-domain generalization. The pivot features are chosen to be both predictive of the label and general across domains. Meeting these two criteria requires task-specific heuristics. Furthermore, the pivot features correspond to a small subspace of the feature co-occurrence matrix. We face a tradeoff between the amount of feature co-occurrence information that we can use, and the computational complexity for representation learning and downstream training. 



We avoid this tradeoff by inducing low dimensional feature embeddings directly. We exploit the tendency of many NLP tasks to divide features into \emph{templates}, with exactly one active feature per template~\citep{smith2011linguistic}; this is shown in the center of Figure~\ref{fig:dacomp}. Rather than treating each instance as an undifferentiated bag-of-features, we exploit this template structure to induce \emph{feature embeddings}: dense representations of individual features. Each embedding is selected to help predict the features that fill out the other templates; see Figure~\ref{fig:dacomp}(b). The embeddings for each active feature are then concatenated together across templates, giving a dense representation for the entire instance.



Our feature embeddings are based on the skip-gram model, trained with negative sampling (SGNS; Mikolov et al., 2013)\nocite{mikolov2013efficient}, which is a simple yet efficient method for learning word embeddings. The training objective is to find feature embeddings that are useful for predicting other active features in the instance. For the instance $n \in \{1 \ldots N\}$ and feature template $t \in \{ 1 \ldots T \}$, we denote $f_n(t)$ as the index of the active feature; for example, in the instance shown in Figure~\ref{fig:dacomp}, $f_n(t) = \text{`new'}$ when $t$ indicates the \textsf{previous-word} template. The skip-gram approach induces distinct ``input'' and ``output'' embeddings for each feature, written $\vu_{f_n(t)}$ and $\vv_{f_n(t)}$, respectively. The role of these embeddings can be seen in the negative sampling objective,
\begin{equation}
\ell_n = \frac{1}{T} \sum_{t=1}^T \sum_{t' \neq t}^T \left [ \log \sigma (\mathbf{u}_{f_n (t)}^\top \mathbf{v}_{f_n (t')}) + k \mathbb{E}_{i \sim P_{t'}^{(n)}} \log \sigma (- \mathbf{u}_{f_n (t)}^\top \mathbf{v}_i) \right ]
\label{eq:objective-1}
\end{equation}
where $t$ and $t'$ are feature templates, $k$ is the number of negative samples, $P_{t'}^{(n)}$ is a \emph{noise distribution} for template $t'$, and $\sigma$ is the sigmoid function. 

Feature embeddings can be applied to domain adaptation by learning embeddings of all features on the union of the source and target data sets. The dense feature vector for each instance is obtained by concatenating the feature embeddings for each template. Finally, since it has been shown that nonlinearity is important for generating robust representations~\citep{bengio2013representation}, we follow~\citep{chen2012marginalized} and apply the hyperbolic tangent function to the embeddings. The augmented representation $\mathbf{x}_n^{\text{(aug)}}$ of instance $n$ is the concatenation of the original feature vector and the feature embeddings:
$\mathbf{x}_n^{\text{(aug)}} = \mathbf{x}_n \oplus \tanh [\mathbf{u}_{f_n (1)} \oplus \cdots \oplus \mathbf{u}_{f_n (T)}] $,
where $\oplus$ is vector concatenation.

\section{Experiments}
\label{sec:exp}
We evaluate \sysname~on part-of-speech (POS) tagging: adaptation of English POS tagging from news text to web text, as in the SANCL shared task~\citep{petrov2012overview}.


\subsection{Experiment setup}
\label{sec:impl}

\paragraph{Datasets} We use data from the SANCL shared task~\citep{petrov2012overview}, which contains several web-related corpora (newsgroups, reviews, weblogs, answers, emails) as well as the WSJ portion of OntoNotes corpus~\citep{hovy2006ontonotes}. Following~\citet{schnabel2014flors}, we use sections 02-21 of WSJ for training and section 22 for development, and use 100,000 unlabeled WSJ sentences from 1988 for learning representations. On the web text side, each of the five target domains has an unlabeled training set of 100,000 sentences, along with development and test sets of about 1000 labeled sentences each. 

\paragraph{SVM tagger} While POS tagging is classically treated as a structured prediction problem, we follow~\citet{schnabel2014flors} by taking a classification-based approach.  Specifically, we apply a support vector machine (SVM) classifier, adding dense features from \sysname~(and the alternative representation learning techniques) to the set of basic features. 
We apply sixteen basic feature templates introduced by~\citet{ratnaparkhi1996maximum}. Feature embeddings are learned for all lexical and affix features, yielding a total of thirteen embeddings per instance. 


\paragraph{Competitive systems} We consider two competitive unsupervised domain adaptation methods that require pivot features: Structural Correspondence Learning (SCL; Blitzer et al., 2006)\nocite{blitzer2006domain} and marginalized Denoising Autoencoders (mDA; Chen et al, 2012)\nocite{chen2012marginalized}. We use structured dropout noise for mDA~\citep{yang2014fast}. We also directly compare with \textsc{word2vec}\footnote{https://code.google.com/p/word2vec/} word embeddings, and with a baseline approach in which we simply train on the source domain data using the surface features, and then test on the target domain. Aside from our own implemented methods, we compare against published results of FLORS~\citep{schnabel2014flors}, which uses distributional features for domain adaptation. We also republish the results of Schnabel and Schutze using the Stanford POS Tagger, a maximum entropy Markov model (MEMM) tagger.


\paragraph{Parameter tuning}
All the hyper-parameters are tuned on development data. Following~\citet{blitzer2006domain}, we consider 6918 pivot features that appear more than 50 times in all the domains for SCL and mDA. The best parameters for SCL are dimensionality $K = 50$ and rescale factor $\alpha = 5$.  For both \sysname~and word2vec, the best embedding size is $100$ and the best number of negative samples is $5$. 
The noise distribution $P_t^{(n)}$ is simply the unigram probability of each feature in the template $t$. \citet{mikolov2013distributed} argue for exponentiating the unigram distribution, but we find it makes little difference here. The window size of word embeddings is set as $5$.

\begin{table*}
\centering
\small
\begin{tabular}{l l l l l l l l}
    \toprule
    Target & baseline & MEMM & SCL & mDA & word2vec & FLORS & \sysname \\ \midrule
    \textsc{newsgroups} & 88.56 & 89.11 & 89.33 & 89.87 & 89.70 & 90.86 & \textbf{91.26} \\
    \textsc{reviews}    & 91.02 & 91.43 & 91.53 & 91.96 & 91.70 & \textbf{92.95} & 92.82 \\
   \textsc{weblogs}    & 93.67 & 94.15 & 94.28 & 94.18 & 94.17 & 94.71 & \textbf{94.95} \\
    \textsc{answers}    & 89.05 & 88.92 & 89.56 & 90.06 & 89.83 & 90.30 & \textbf{90.69} \\
    \textsc{emails}     & 88.12 & 88.68 & 88.42 & 88.71 & 88.51 & 89.44 & \textbf{89.72} \\ 
    \bottomrule
\end{tabular}
\caption{Accuracy results for adaptation from WSJ to Web Text on SANCL dev set. }
\label{tab:sancldev}
\end{table*}

\begin{table*}
\centering
\small
\begin{tabular}{l l l l l l l l l}
    \toprule
    Target & baseline & MEMM & SCL & mDA & word2vec & FLORS & \sysname \\ \midrule
    \textsc{newsgroups} & 91.02 & 91.25 & 91.51 & 91.83 & 91.35 & 92.41 & \textbf{92.60} \\
    \textsc{reviews}    & 89.79 & 90.30 & 90.29 & 90.95 & 90.87 & \textbf{92.25} & 92.15 \\
    \textsc{weblogs}    & 91.85 & 92.32 & 92.32 & 92.39 & 92.42 & 93.14 & \textbf{93.43} \\
    \textsc{answers}    & 89.52 & 89.74 & 90.04 & 90.61 & 90.48 & 91.17 & \textbf{91.35} \\
    \textsc{emails}     & 87.45 & 87.77 & 88.04 & 88.11 & 88.28 & 88.67 & \textbf{89.02} \\ 
    \bottomrule
\end{tabular}
\caption{Accuracy results for adaptation from WSJ to Web Text on SANCL test set.}
\label{tab:sancltest}
\end{table*}

\subsection{Results} 
As shown in Table~\ref{tab:sancldev} and~\ref{tab:sancltest}, \sysname~outperforms competitive systems on all target domains except \textsc{review}, where FLORS performs slightly better. FLORS uses more basic features than \sysname; these features could in principle be combined with feature embeddings for better performance. Compared with the other representation learning approaches, \sysname~is roughly 1\% better on average, corresponding to an error reduction of 10\%. Its training time is approximately 70 minutes on a 24-core machine, using an implementation based on gensim.\footnote{\url{http://radimrehurek.com/gensim/}} This is slightly faster than SCL, although slower than mDA with structured dropout noise.

%
%
%

\section{Related Work}
\label{sec:related}
Representation learning approaches to domain adaptation seek for cross-domain representations, which were first induced via auxiliary prediction problems~\citep{ando2005framework}, such as the prediction of \emph{pivot features}~\citep{blitzer2006domain}. In these approaches, as well as in later work on denoising autoencoders~\citep{chen2012marginalized}, the key mechanism is to learn a function to predict a subset of features for each instance, based on other features of the instance. 

Word embeddings can be viewed as special case of representation learning, where the goal is to learn representations for each word, and then to supply these representations in place of lexical features~\citep{turian2010word}. 

\section{Conclusion}
\label{sec:con}

Feature embeddings can be used for domain adaptation in any problem involving feature templates. They offer strong performance, avoid practical drawbacks of alternative representation learning approaches, and are easy to learn using existing word embedding methods. 

\bibliography{cite-strings,cites,cite-definitions}
\bibliographystyle{iclr2015}

\end{document}